\documentclass[runningheads,a4paper]{llncs}

\usepackage{amssymb}
\usepackage{graphicx}

\usepackage{url}
\urldef{\mailsa}\path|{alfred.hofmann, ursula.barth, ingrid.haas, frank.holzwarth,|
\urldef{\mailsb}\path|anna.kramer, leonie.kunz, christine.reiss, nicole.sator,|
\urldef{\mailsc}\path|erika.siebert-cole, peter.strasser, lncs}@springer.com|    
\newcommand{\keywords}[1]{\par\addvspace\baselineskip
\noindent\keywordname\enspace\ignorespaces#1}

\begin{document}

\mainmatter  % start of an individual contribution

\title{Improving Siamese Networks for One Shot Learning using Kernel Based Activation functions}

\titlerunning{Lecture Notes in Computer Science: Authors' Instructions}

\author{Shruti Jadon
\and Aditya Arcot Srinivasan}
% \and Ingrid Haas\and Frank Holzwarth\and\\
% Anna Kramer\and Leonie Kunz\and Christine Rei\ss\and\\
% Nicole Sator\and Erika Siebert-Cole\and Peter Stra\ss er}
%
% \authorrunning{Lecture Notes in Computer Science: Authors' Instructions}

\institute{Department of Computer Science\\
University of Massachusetts Amherst\\
Amherst, MA \\
% Springer-Verlag, Computer Science Editorial,\\
% Tiergartenstr. 17, 69121 Heidelberg, Germany\\
% \mailsa\\ 
% \mailsb\\ 
\url{sjadon@umass.edu, asrinivasan@cs.umass.edu}
}


\toctitle{Lecture Notes in Computer Science}
\tocauthor{Authors' Instructions}
\maketitle

\begin{abstract}
The lack of a large amount of training data has always been the constraining factor in solving a lot of problems in machine learning, making One Shot Learning one of the most intriguing ideas in machine learning. It aims to learn information about object categories from one, or only a few training examples. This process of learning in deep learning is usually accomplished by proper objective function, i.e; loss function and embeddings extraction i.e; architecture. In this paper,
we discussed about metrics based deep learning architectures for one shot learning such as Siamese neural networks \cite{koch2015siamese} and present a method to improve on their accuracy using \emph{Kafnets} (kernel-based non-parametric activation functions for neural networks) \cite{scardapane2017kafnets} by learning proper embeddings with relatively less number of epochs. Using kernel activation functions, we are able to achieve strong results which exceed those of ReLU based deep learning models in terms of embeddings structure, loss convergence, and accuracy. The project code with results can be found at \url{ https://github.com/shruti-jadon/Siamese-Network-for-One-shot-Learning}.

% The abstract should summarize the contents of the paper and should
% contain at least 70 and at most 150 words. It should be written using the
% \emph{abstract} environment.
\end{abstract}

\section{Introduction}
Humans learn new things with a very small set of examples – e.g. a child can generalize the concept of a "Dog" from a single picture but a machine learning system needs a lot of examples to learn its features. In particular, when presented with stimuli, people seem to be able to understand new concepts quickly and then recognize variations on these concepts in future percepts \cite{lake2011one}. Machine learning as a field has been highly successful at a variety of tasks such as classification, web search, image and speech recognition. Often times however, these models do not do very well in the regime of low data. This is the primary motivation behind One Shot Learning; to train a model with fewer examples but generalize to unfamiliar categories without extensive retraining.

Deep learning has played an important role in the advancement of machine learning, but it also requires large datasets. Different techniques such as regularization reduces overfitting in low data regimes, but do not solve the inherent problem that comes with fewer training examples. Furthermore, the large size of datasets leads to slow learning, requiring many weight updates using stochastic gradient descent. This is mostly due to the parametric aspect of the model, in which training examples need to be slowly learned by the model into its parameters. In contrast, many known non-parametric models like nearest neighbors do not require any training but performance depends on a sometimes arbitrarily chosen distance metric like the L2 distance[1].

One-shot learning is an object categorization problem in computer vision. Whereas most machine learning based object categorization algorithms require training on hundreds or thousands of images and very large datasets \cite{Jadon2017VideoSU}, one-shot learning aims to learn information about object categories from one, or only a few, training images \cite{fei2006one}. 

One way of addressing problems in One Shot learning is to develop specific features relevant to the domain of the problem; features that possess discriminative properties particular to a given target task. However, the problem with this approach is the lack of generalization that comes along with making assumptions about the structure of the input data. In this paper, we make use of an approach similar to \cite{koch2015siamese} \cite{jadon2019hands} while simultaneously evaluating different activation functions \cite{jadon2018introduction} that may be better suited to this task. The overall strategy we apply is two fold; train a discriminative deep learning model on a collection of data with similar/dissimilar pairs. Then, using the learned feature mappings, we can evaluate new categories.

Since One  Shot  Learning  focuses  on  models  which  have  a  nonparametric  approach  of  evaluation,  we  came  across  Kafnets \cite{scardapane2017kafnets} (kernel  based  non-parametric  activation functions)  that  have shown initial promise in  this  domain  of  training  neural networks  using  different  forms  of  activation  functions; so as to increase  non-linearity,  therefore  decreasing  the  number  of layers, and increasing the accuracy in a lot of cases. This paper has  proposed  two  activation  functions KAF and KAF2D,  and  focuses  on  their nature of continuity and differentiability. We have implemented these activations and compared their effectiveness against traditional ones when used in the context of One Shot learning.
\subsection{Related Work}
The research in one shot learning has recently caught attention of the machine learning community. The work resulting in the best accuracy for the image classification problem dates back to the 2000's by \cite{fei2006one}. The authors
have developed a variational bayesian framework \cite{fei2006one} \cite{jadon2019hands}for one
shot image classification using the premise that a previously
learned class can help in forecasting a future one.

\cite{lake2013one} tackled the problem of character recognition by proposing a method called Hierarchical Bayesian Program Learning. In \cite{lake2011one} and \cite{lake2013one}, the authors present an approach where an image is deconstructed into several smaller pieces to ascertain an explanation for the structure of pixels. However, the joint parameter space being very large lead to inference becoming intractable.

There have also been other methods that approach the problem of One Shot Learning. \cite{wu2012one} tackle path planning as a one shot learning problem for robotic actuation. \cite{maas2009one} use Bayesian networks on the Ellis Island passenger data to infer attributes. \cite{lake2014one} use a generative Hidden Markov Model along with a Bayesian inference algorithm to try and identify unseen words in a speech recognition paradigm. \cite{bertinetto2016learning} predicts
the parameters of a neural network from a single exemplar image. The network
that effectively learns to learn, generalizing across tasks defined by different exemplars.

A different approach to one-shot learning is to learn an embedding space, which is typically done with a siamese network \cite{bromley1994signature}. Given an exemplar of a novel category, classification is performed in the embedding space by a simple rule such as nearest-neighbor. Training is usually performed by classifying pairs according to distance \cite{fan2014learning}.

Another technique that looks at the problem of One Shot Learning is by use of matching networks or bi-directional LSTMs \cite{vinyals2016matching}. As mentioned before, non parametric alternatives like the Nearest Neighbours model choose an arbitrary distance function. The authors solve this problem by formulating a loss function that encompasses in training a nearest neighbour like model end to end. In the image classification task, the generated output label $\hat{y}$ for a test example $\hat{x}$ is computed very similar to what you might see in Nearest Neighbors algorithm. The method progresses by embedding both the training examples as well as given test example $\hat{x}$, compute a cosine similarity based metric as the "match", and then pass that through a softmax to get normalized mixing weights to generate a label. The embedding process for the training examples make use of a bidirectional LSTM over the examples. For the test examples,  is a an LSTM that processes for a fixed amount (K time steps) and at each point also attends over the examples in the training set. The encoding is the last hidden state of the LSTM. The paper also benchmarks various approaches to one shot learning could be used a reference for our results.

The approach that has been recently explored is the use of Deep Siamese Networks which we borrow from heavily \cite{koch2015siamese}. Convolutional neural networks have achieved exceptional results in many large-scale computer vision applications, particularly in image recognition tasks. Several factors make convolutional networks especially appealing. Local connectivity can greatly reduce the number of parameters in the model, which inherently provides some form of built-in regularization, although convolutional layers are computationally more expensive than standard non-linearities. Also, the convolution operation used in these networks has a direct filtering interpretation, where each feature map is convolved against input features to identify patterns as groupings of pixels. Thus, the outputs of each convolutional layer correspond to important spatial features in the original input space and offer some robustness to simple transforms. Further we use a contrastive loss function as defined in \cite{chopra2005learning}. The objective of the siamese architecture is not to classify input images, but to differentiate between them. So, a classification loss function (such as cross entropy) would not be the best fit. Instead, this architecture is better suited to use a contrastive function. Intuitively, this function just evaluates how well the network is distinguishing a given pair of images.

As One Shot Learning focuses on models which have a non parametric approach of evaluation, we came across work \cite{scardapane2017kafnets} on Kafnets (kernel based non-parametric activation functions) that has worked in this domain of training neural networks using different forms of activation functions. \cite{scardapane2017kafnets} introduce a novel family of flexible activation functions that are based on an inexpensive kernel expansion at every neuron. Leveraging over several properties of kernel-based models, the authors propose multiple variations for designing and initializing these kernel activation functions (KAFs), including a multidimensional scheme allowing to non linearly combine information from different paths in the network. The resulting KAFs can approximate any mapping defined over a subset of the real line, either convex or nonconvex. Furthermore, they are smooth over their entire domain, linear in their parameters, and they can be regularized using any known scheme. In this paper, we focus on two activation functions, KAF and KAF2D and the effects of implementing them in a siamese architecture for One Shot Learning.

\subsection{Dataset}
For this experiment, we have used three main datasets: MNIST \cite{lecun1998mnist}, AT\&T Database of Faces \cite{attdataset} (formerly 'The ORL Database of Faces'), and  Omniglot dataset \cite{lake2015human}. All of the above datasets are available freely online and did not require any form of preprocessing.
We chose MNIST because we first wanted to test our models with images with less information. Then to showcase generalization of our approach we chose dataset with more features to extract from images and decided to work with the AT\&T Database of Faces. The MNIST dataset is a collective dataset of 0-9 handwritten digits by various humans, It consists of training-set of 60,000 examples, and a test set of 10,000 examples, but as we are experimenting on One-Shot Learning approach, we have randomly sampled and chose only 500 images of available dataset.
% \begin{figure}
% \centering
%   \includegraphics[scale=0.2]{mnist_sample.png}
%   \caption{Sample MNIST Dataset}
% \end{figure}
Similar to MNIST, The AT\&T Database of Faces consists of ten different images of each of 40 different human subjects. For certain subjects, Images consists of variation in times, light, and expressions.
% \begin{figure}
% \centering
%   \includegraphics[scale=0.8]{at_sample.jpg}
%   \caption{Sample AT\&T Face Dataset}
% \end{figure}

\begin{figure}[htp]

\centering
\includegraphics[width=.2\textwidth]{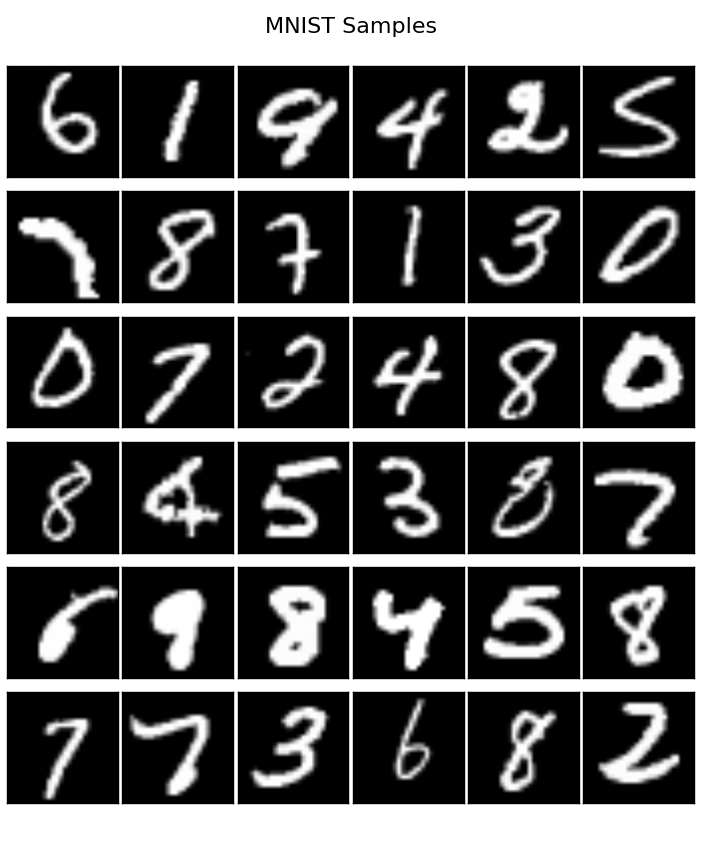}\hfill
\includegraphics[width=.3\textwidth]{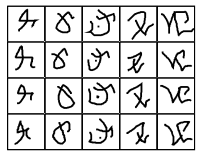}\hfill
\includegraphics[width=.4\textwidth]{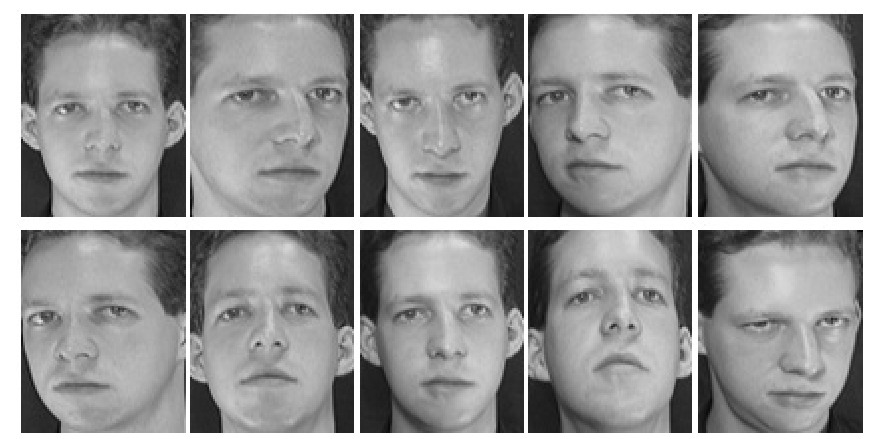}

\caption{Sample MNIST, Omniglot, and AT\&T Dataset}
\label{fig:figure3}

\end{figure}
Omniglot is a dataset by \cite{lake2015human} is one of the most popular dataset used for one-shot learning. It is specially designed to compare and contrast the learning abilities of humans and machines. This dataset consists of handwritten
characters of 50 languages (alphabets) with 1623 total characters. There are only 20 samples for each character, each drawn by a distinct individual. The dataset is divided into 2 sets: background set and evaluation set. Background set contains 30 alphabets (964 characters) and is used to train the model whereas, the remaining 20 alphabets are for pure evaluation purposes only. 

\section{Methodology}
For this work, we first implemented basic architecture of Siamese Networks and Matching Networks using ReLU Activation function, and compared their learning abilities(Embeddings, and Optimization Ability) in comparison to Kernel Activation Functions. Siamese network\cite{koch2015siamese} is an architecture with two parallel layers. In this architecture, instead of a model learning to classify its inputs using classification loss functions, the model learns to differentiate between two given inputs. It compares two inputs based on a similarity metric, and checks whether they are same or not. Similar to any deep learning architecture, a Siamese network also has two phases: Training and Testing Phase. But usually for a One-shot learning approach (as we won't have a lot of data points) we train the model architecture on a different dataset and test it for our less amount of dataset. To put in simpler terms, we learn image embeddings using a supervised metric-based approach with Siamese neural networks, then reuse that network's features for One-shot learning without fine-tuning or retraining.
\begin{figure}
\centering
  \includegraphics[scale=0.5]{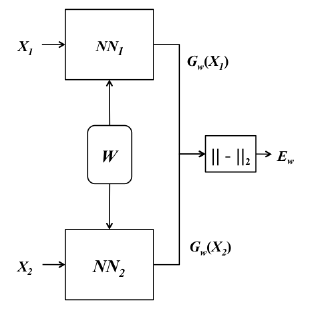}
  \caption{Sample Siamese Networks Architecture}
\end{figure}

In this work, we have compared embeddings and optimization of Siamese Network Architecture with ReLU Activation Function and Kernel Activation Functions. Through experimentation, we made 2 keen observations: 
\begin{enumerate}
    \item Embeddings learned using Kernel Activation functions were clearly separated as compared to in ReLU Function.
    \item As Learning Capacity increased due to Kernel Activation Functions, it led to faster learning i.e; better accuracy within few number of epochs.
\end{enumerate}

One challenging task involved in this experiment is implementation of activation functions. For this, We took help of kernel\cite{scardapane2017kafnets} implementation of two Activation Functions, called KAF and KAF2D. KAF(Kernel Activation Function) Specifically, each activation function is modelled in terms of a gaussian kernel expansion over D-dimension terms as: \\
$ g(s)= \sum_{i=1}^{D}\alpha_i\kappa(s,d_i) $ where, $\left \{ \alpha_i \right \}_{i=1:D}$
are the mixing coefficients, $\left \{ d_i \right \}_{i=1:D}$ are the called the dictionary elements, and $\kappa\left ( .,. \right ):\mathbb{R}\rightarrow \mathbb{R}$  is 1D kernel function.

In Simpler terms, we are trying to project every layer output in a form of gaussian distribution and add them across D iterations. But If we observe in general Stochastic Gradient Training settings, D will increase as number of training iterations increases. Therefore, we are fixing D to a number, and just varying mixing coefficients ($\alpha$) of kernel functions. This has benefit that the resulting function is linear in its adaptable parameters, and can be implemented for a mini-batch of training data. Stochastic Gradient descent works properly if function is convex in nature, in matrix terms, loss matrix need to abide by the positive semi-definitive property. Kernel activation function also maintains the convexity of the function across layers, whereas ReLU causes function to loose its convexity. For kernel function to respect the positive semi-definiteness property, it can be of any possible $\alpha_i$ and $d_i$ in:
\[\sum_{i=1}^{D}\sum_{j=1}^{D} \alpha_i \alpha_j \kappa(d_i,d_j)\geq 0\]
As we are using 1D Gaussian kernel $\kappa(s,d_i)=exp{\left \{ -\gamma(s-d_i)^2 \right \}}$ where $\gamma \epsilon  \mathbb{R}$ is called the kernel bandwidth.

This gives us a proper derivatives for back-propagation as seen below:
\[\frac{\partial g(s)}{\partial \alpha_i} = \kappa (s,d_i)\]
\[\frac{\partial g(s)}{\partial s} =\sum_{i=1}^{D}\alpha_i \frac{\partial \kappa(s,d_i)}{\partial s}\]
%On the selection of the kernel bandwidth, which is a crucial step.

\cite{scardapane2017kafnets} also considers a two-dimensional variant of the
proposed KAF, denoted as 2D-KAF. Roughly speaking, the 2D-KAF acts on a pair of activation values, instead of a single one, and learns a two dimensional function to combine them. It can be seen as a generalization of a two-dimensional max-out neuron, which is instead constrained to output the maximum value among the two inputs. The equation is given as:
\[g(s)= \sum_{i=1}^{D^2}\alpha_i\kappa(s,d_i)\]
here, $\kappa(s,d_i)=exp\left \{ -\gamma \left \| s-d_i \right \|_2^2 \right \}$ is a 2D Gaussian Kernel.

\subsection {Siamese Network Architecture}
We have experimented on Siamese Networks with 2 Datasets: MNIST and AT\&T Dataset. For MNIST, we have coded a very simple 2 convolutional layer architecture: \\
Layer 1 : Conv1 and Conv2\\
Conv1: 1X20 followed by maxpooling \\
Conv2: 20X50 followed by maxpooling \\
Layer 2 : Fully Connected network with Activation Function. (50X500) \\
Layer 3 : Linear Layer (500X2) \\

As Face features are complex, For AT\&T Face Dataset, we have implemented dense layered Architecture: \\
Layer 1 : Conv1, Conv2, and Conv3 \\
Conv1: 1X4 followed by Activation Function and Max Pooling. \\
Conv2: 4X4 followed by Activation Function and Max Pooling. \\
Conv3: 8X8 followed by Activation Function and Max Pooling. \\
Layer 2 :  Fully Connected network with Activation Function (100X100X8). \\
Layer 3 : Linear Layer with Activation Function. (500X250). \\
Layer 4 : Linear Layer with Activation Function. (250X5). \\
We have used Contrastive Loss function in both cases:
\[(1-Y)\frac{1}{2}D_w^2 + (Y) \frac{1}{2}\left \{ max(0,m-D_w) \right \}^2 , and \]

\[D_w=\sqrt{\left \{ G_w(X1)-G_w(X2)\right \}^2}\]

where $G_w$ is the output of one of the sister networks. X1 and X2 is the input data pair. \\
For training, we have set learning rate to 0.0005, and used the Adam Optimizer. We have also experimented on matching networks\cite{vinyals2016matching} using 1-D and 2-D Kernel Functions. For Matching Networks we created 2 convoluted embeddings extraction layer, followed by 3 Fully connected linear layers and Bidirectional LSTMs for full contextual embeddings extraction. 
\section{Experiments and Results}
For Analysis of embeddings output we compared intercluster vs intracluster distance of the features projection of different inputs, and we used silhouette score. We trained Siamese Network Architecture on the MNIST Dataset \cite{jadon2019hands}, with different activation functions \cite{jadon2018introduction}, for 10 and 50 epochs. We have observed that the clustering score (silhouette score) in KAF2D was best, followed by KAF and RELU as displayed in Table 1. Silhouette Score ranges from (-1 to 1), where close to 1 proves that the clusters obtained are good. There are 2 main reasons behind imporved performace:
\begin{enumerate}
    \item For one shot non-paramteric activation functions can be of added value, as we have less amount of data. Though, ReLU is simple traditional non-paramteric function, but it doesn't add the capacity to architecture, rather it relies on added neural network layers. Whereas KAF, and KAF2D are gaussian kernel based, therefore enabling layers to capture better features of images.
    \item Gaussian Distribution is considered as best approximator when distribution is unknown, as stated by Central Limit Theorem. Therefore, for cases of less images, approximating distribution of data after each layer, will lead to better convergence.
\end{enumerate}
\begin{figure}
\centering
  \includegraphics[scale=0.11]{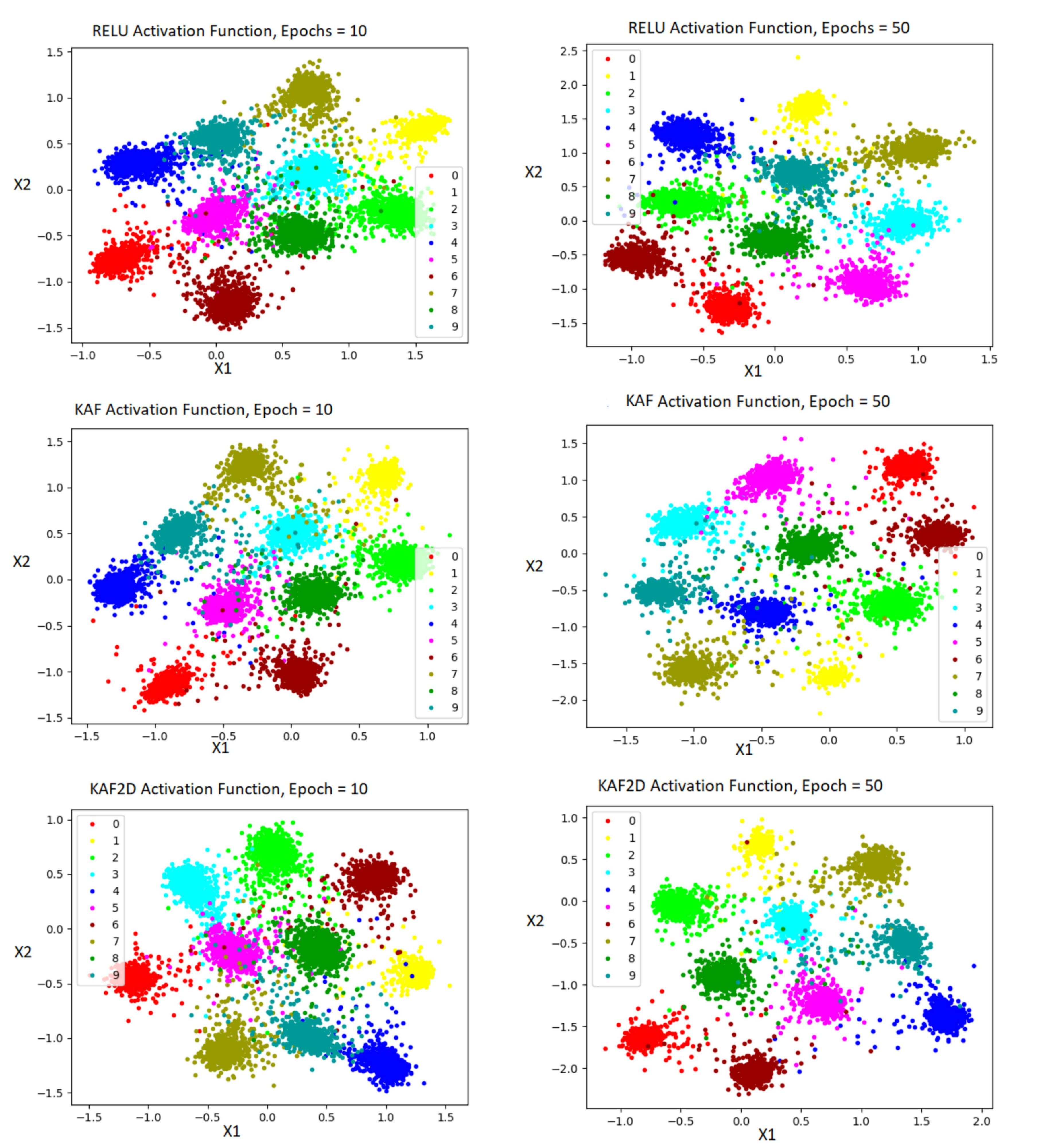}
  \caption{Embeddings with different Activation Function for different Epochs (MNIST Dataset)}
\end{figure}
We have also observed the training loss curve for all the cases, and see that with KAF, and KAF2D there were more fluctuations than RELU, refer figure in Supplemetary Material.

% \begin{figure}[h!]
% \centering
%   \includegraphics[scale=0.10]{Images/Embedding_KAF2D.jpg}
%   \caption{Embeddings with KAF2D Activation Function for different Epochs}
% \end{figure}

We also experimented with the same architecture on the AT\&T Face Similarity dataset. As the output from the Siamese network, we obtained five dimensional embeddings of the images in a plane; we then calculated pairwise distance which was used the metric to measure similarity.
Similar to the MNIST experiment, we ran it with KAF, KAF2D, and RELU activation function, and observed that we were able to increase the closeness of it, using KAF and KAF2D. We also observed the behavior of training loss curve with different functions. What we observed is, that for RELU, it converged much faster, which could be reason of its efficiency. Whereas KAF, and KAF2D were fluctuating in the beginning, but converged to a lower value of loss at the end.
\begin{figure}
\centering
  \includegraphics[scale=0.07]{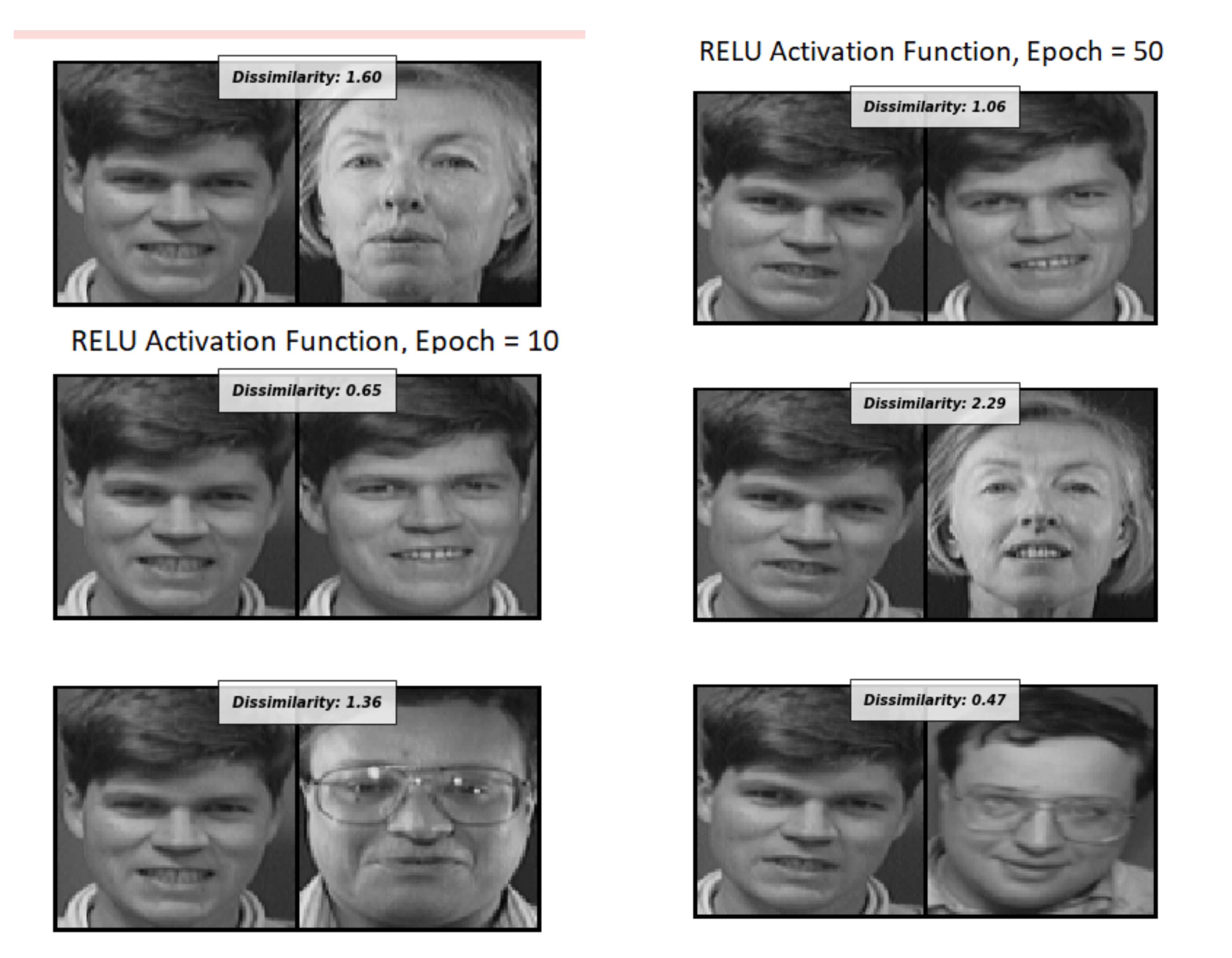}
  \caption{Similarity Scores for Faces with RELU Activation Function (AT\&T Dataset)}
\end{figure}
\begin{figure}
\centering
  \includegraphics[scale=0.07]{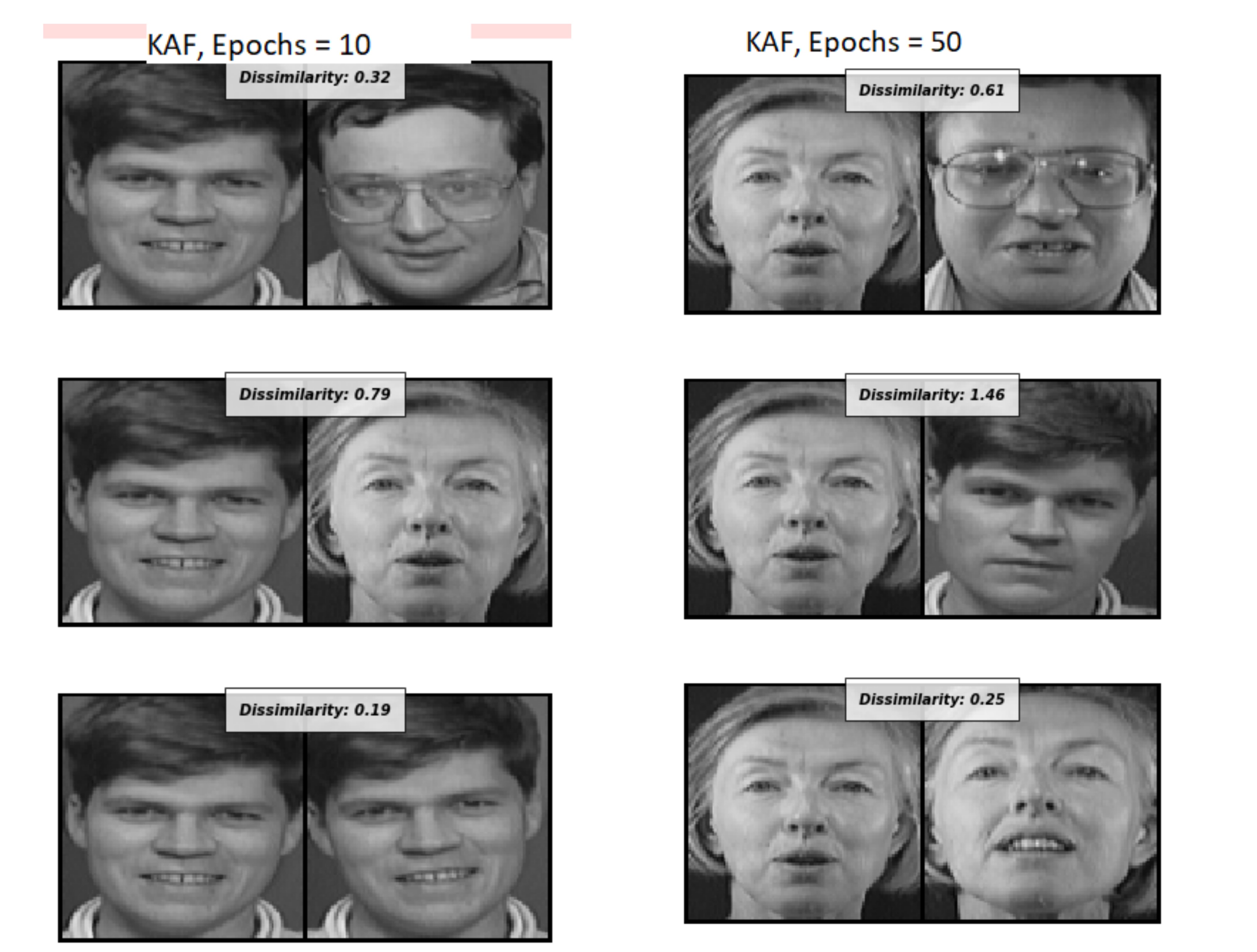}
  \caption{Similarity Scores for Faces with KAF Activation Function (AT\&T Dataset)}
\end{figure}

\begin{table}
\caption{Silhouette Scores obtained for clusters in test set, 50 Epochs [MNIST Dataset]}
\label{sample-table1}
\begin{center}
\begin{tabular}{ll}
\multicolumn{1}{c}{\bf Activation Function}  & \multicolumn{1}{c}{\bf Silhouette Score}
\\ \hline \\
RELU         & 0.7766 \\
KAF             & 0.8052 \\
KAF2D             & 0.81641 \\
\end{tabular}
\end{center}
\end{table}

\begin{table}
\caption{Accuracies for Matching Networks on Omniglot Dataset}
\label{sample-table2}
\begin{center}
\begin{tabular}{ll}
\multicolumn{1}{c}{\bf Activation Function}  & \multicolumn{1}{c}{\bf Accuracy}
\\ \hline \\
RELU         & 80.625\% \\
KAF             & 85.06\% \\
KAF2D             & 89.27\% \\
\end{tabular}
\end{center}
\end{table}

% \subsection{Matching Networks Experimets}
As a final experiment, we replicated the architecture of Matching Networks as in \cite{vinyals2016matching} with the KAF and KAF2D activation functions on the Omniglot Dataset. we ran it for 1600 Epochs with the results summarized in Table 2.

\section{Discussion and Conclusions}
In this paper, we presented a method to improve metrics based one-shot learning approaches using kernel based activation functions \cite{scardapane2017kafnets}. We have outlined our results comparing the performance of our networks to existing ReLU based Architectures. After running experiments, we observe certain behavior related to activation functions as applied to the One Shot learning task:
\begin{enumerate}
  \item We obtained better clusters (closely aligned) with KAF2D, followed by KAF and RELU, for MNIST Dataset.
  \item The Training Loss Curve for AT\&T Face Dataset converged faster when using RELU. When using KAF and KAF2D as activation functions, the loss fluctuated a bit in the beginning but provided a lower loss value at the end.
   \item KAF takes around twice the training time of RELU activation functions.
  \item KAF2D takes around five times the training time of RELU activation functions.
  \item The Results for Matching Networks Architecture proved to be promising using Kernel based activation functions.
\end{enumerate} 
 We conclude that though new Activation Functions gives better accuracy in matching networks, better similarity distance in AT\&T dataset, and better intra cluster scores for MNIST in less number of epochs, but it take a lot more time to converge as compared to RELU.\\
Learning proper representations is an crucial part of any deep learning architecture, especially in case of one shot learning when we have less amount of data. To learn the ability to learn, we need to improve out method of information extraction. In above work, we proposed one method of data approximation using gaussian distribution activation function. For metrics based one shot learning, we can at least say that we need to learn proper approximation functions instead of adding more layers, as adding layers will led to underfitting/overfitting model considering less amount of data.

\medskip

\small

% \nocite{*} % to test all bib entrys
\bibliographystyle{unsrt}
% \begin{thebibliography}{4}
\bibliography{egbib.bib} % file mwe.bib

\section{Supplementary Material}
\begin{figure}
\centering
  \includegraphics[scale=0.09]{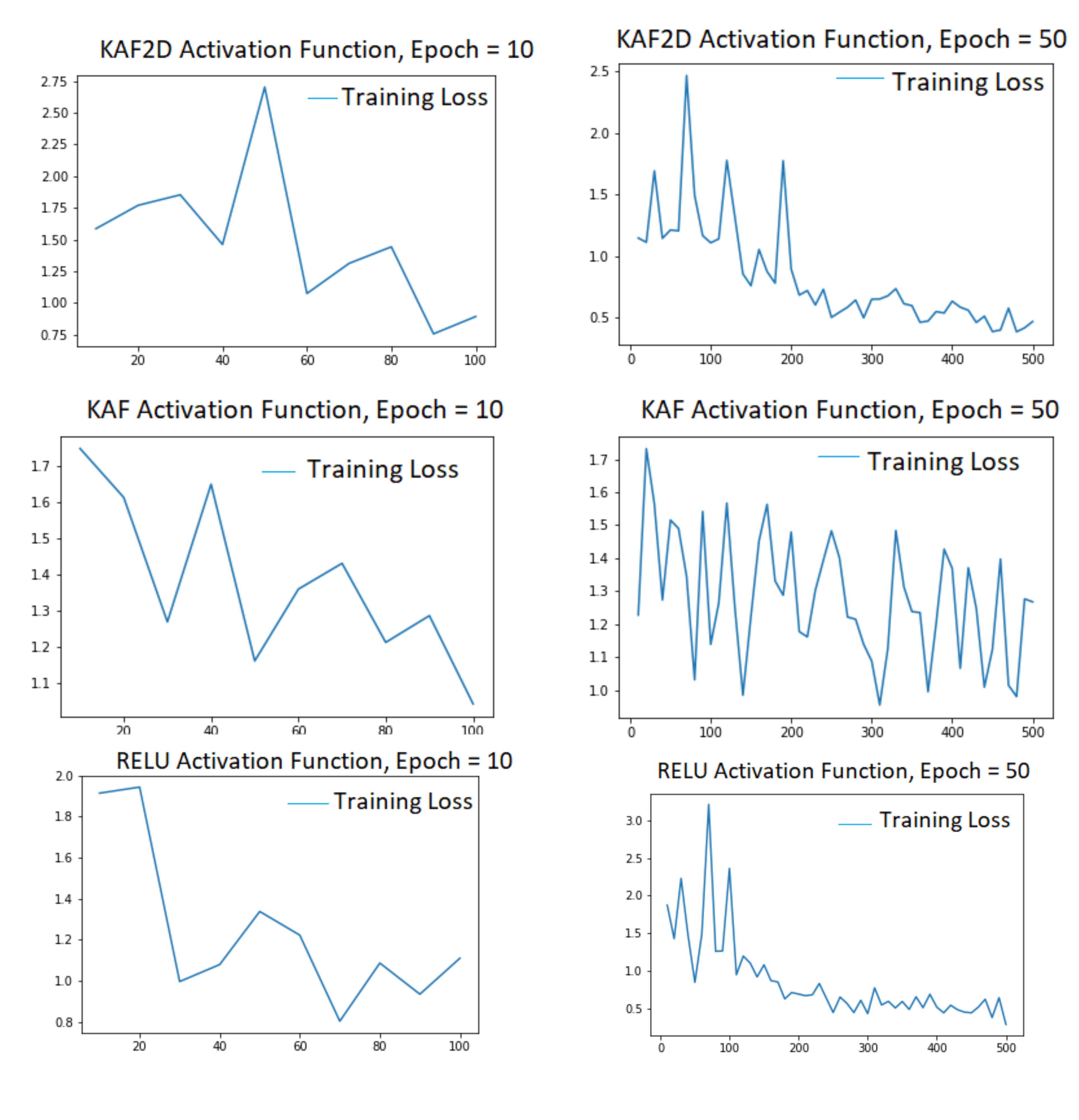}
  \caption{Training Loss for Different Epochs AT\&T Dataset}
\end{figure}
\begin{figure}
\centering
  \includegraphics[scale=0.085]{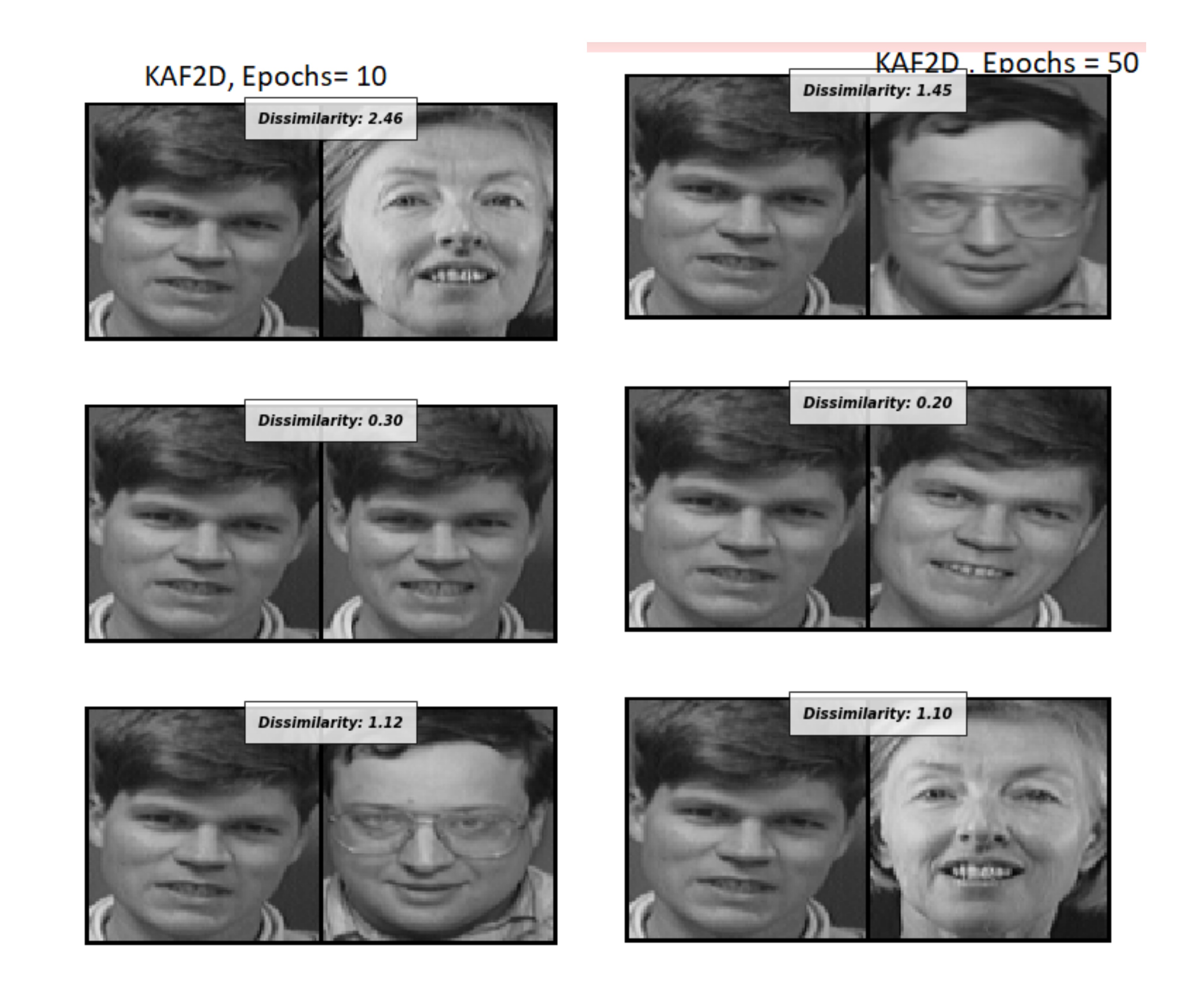}
  \caption{Similarity Scores for Faces with KAF2D Activation Function [AT\&T Dataset]}
\end{figure}
\begin{figure}
\centering
  \includegraphics[scale=0.12]{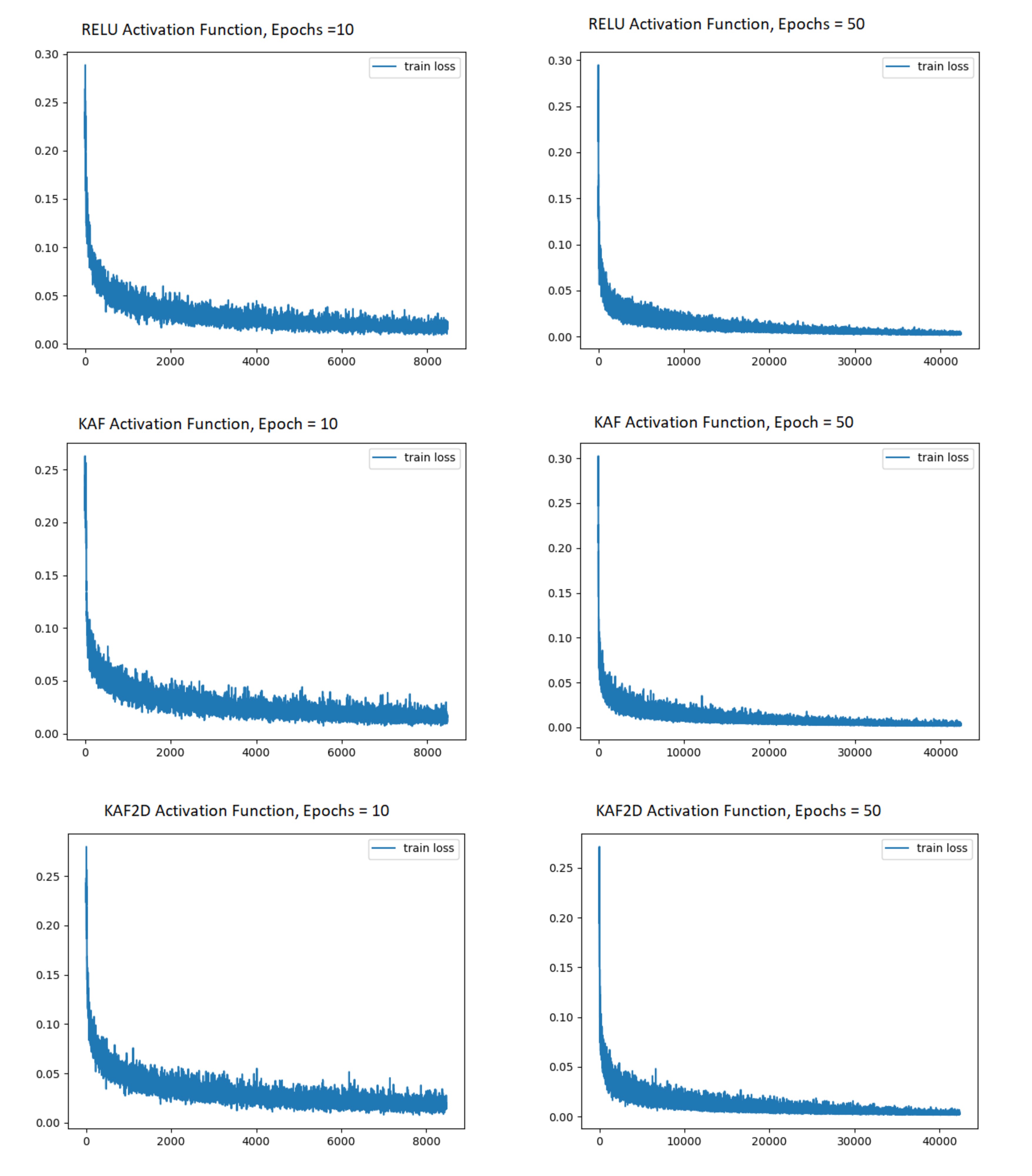}
  \caption{Training Loss for MNIST dataset with different Epochs and Activation Functions.}
\end{figure}
\end{document}